\documentclass{article}

\usepackage[preprint]{neurips_2026}
\makeatletter
\AtEndDocument{%
  \immediate\write\@auxout{\string\bibstyle{plainnat}}%
  \immediate\write\@auxout{\string\bibdata{dummy}}%
}
\makeatother

\usepackage[utf8]{inputenc}
\usepackage[T1]{fontenc}
\usepackage[colorlinks=true,linkcolor=blue,citecolor=blue,urlcolor=blue]{hyperref}
\usepackage{url}
\usepackage{booktabs}
\usepackage{amsfonts}
\usepackage{amsmath}
\usepackage{amssymb}
\usepackage{nicefrac}
\usepackage{microtype}
\usepackage{xcolor}
\usepackage{graphicx}
\usepackage{float}
\graphicspath{{figures/}}
\usepackage{subcaption}
\usepackage{multirow}
\usepackage{wrapfig}
\usepackage{enumitem}

\newcommand{\ours}{AGWM}
\newcommand{\sconly}{\textbf{SC-Only}}

\newcommand{\nval}[1]{#1}

\title{AGWM: Affordance-Grounded World Models for Environments with Compositional Prerequisites}

\makeatletter
\renewcommand{\@bottomtitlebar}{%
  \vskip 0.10in
  \vskip -\parskip
  \vskip 0.04in%
}
\renewcommand{\@maketitle}{%
  \vbox{%
    \hsize\textwidth
    \linewidth\hsize
    \vskip 0.1in
    \@toptitlebar
    \centering
    {\LARGE\bf \@title\par}
    \@bottomtitlebar
    \begin{tabular}[t]{c}\@author\end{tabular}%
    \vskip 0.15in \@minus 0.1in
  }
}
\makeatother

\author{%
  \normalfont Qinshi Zhang$^{1}$ \quad Weipeng Deng$^{2}$ \quad Zhihan Jiang$^{3}$ \quad Jiaming Qu$^{4}$\thanks{Work does not relate to the author's position at Amazon.} \\[2pt]
  Qianren Li$^{5}$ \quad Weitao Xu$^{5}$ \quad Ray LC$^{5}$ \\[3pt]
  {\small $^{1}$University of California, San Diego \quad
  $^{2}$University of Hong Kong \quad
  $^{3}$Columbia University} \\[1pt]
  {\small $^{4}$Amazon \quad
  $^{5}$City University of Hong Kong} \\[2pt]
  {\small \texttt{qiz065@ucsd.edu} \quad \texttt{vincentdengp@icloud.com} \quad \texttt{zj2445@cumc.columbia.edu} \quad \texttt{qjiaming@amazon.com}} \\[1pt]
  {\small \texttt{qianrenli2@cityu.edu.hk} \quad \texttt{weitaoxu@cityu.edu.hk} \quad \texttt{raylc@cityu.edu.hk}}
}

\begin{document}

\maketitle

\vspace{-0.75cm}
\begin{figure}[!h]
  \centering
  \includegraphics[width=\dimexpr\linewidth-6pc\relax]{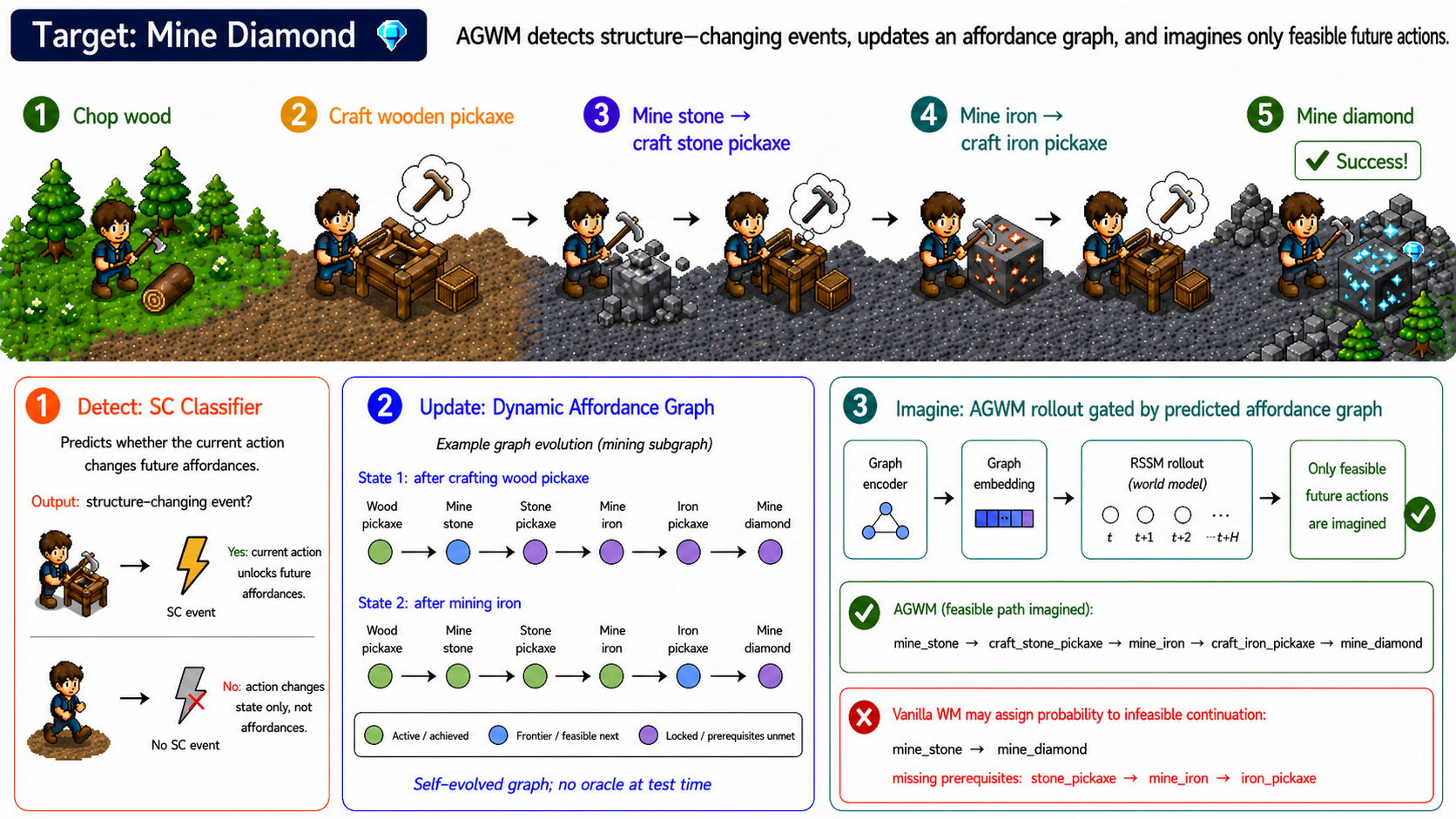}
  \caption{\textbf{AGWM overview.} \textbf{Top:} The agent traverses a four-tier tech tree; SC events (colored markers) progressively expand the applicable action set. \textbf{Bottom:} \ours{} operates in three stages: (1)~\textbf{Detect} SC events via the SC Classifier; (2)~\textbf{Update} the Dynamic Affordance Graph to track active (green), frontier (blue), and locked (purple) capabilities without oracle input; (3)~\textbf{Imagine} by gating RSSM rollouts with the graph embedding, so only feasible action sequences are considered.}
  \label{fig:agwm_overview}
\end{figure}
\vspace{-0.2cm}

\begin{abstract}
  In model-based learning, the agent learns behaviors by simulating trajectories based on world model predictions. Standard world models typically learn a stationary transition function that maps states and actions to next states, when an action and an outcome frequently co-occur in training data, the model tends to internalize this correlation as a general causal rule while ignoring action preconditions. In interactive environments, however, agent actions can reshape the future affordance space. At each timestep, an action may becomes executable only after its prerequisites are met, or non-executable when they are destroyed. We term such events \emph{structure-changing events} (SC events). As a result, a conventional world model often fails to determine whether a given action is executable in the current state, especially in multi‑step predictions. Each imagined step is conditioned on an incorrect affordance state, and therefore the prediction error compounds over the rollout horizon. In this paper, We propose AGWM (Affordance‑Grounded World Model), which learns an abstract affordance structure represented as a DAG of prerequisite dependencies to explicitly track the dynamic executability of actions. Experiments on game-based simulated environments demonstrate the effectiveness of our method by achieving lower multi-step prediction error, better generalizataaion to novel configurations, and improved interpretability.
\end{abstract}

\section{Introduction}
\label{sec:intro}

World models encode environment dynamics in neural network parameters, learning statistical associations between states, actions, and their outcomes~\citep{ha2018world, ke2019learning, li2020causal}. Given a learned world model, an agent can roll out imagined trajectories and select the action sequence that maximizes expected return without additional environment interaction~\citep{hafner2023dreamerv3}.
Standard world models often reconstruct observations well, but they do not model affordances, which are the set of actions currently available to the agent~\citep{gibson1979ecological}. They can predict what follows a given action yet cannot reliably determine whether that action is executable in the current state~\citep{khetarpal2020whatcanido}. In open-ended environments such as robotic manipulation and autonomous planning, an agent may confidently commit to action sequences that are physically or logically impossible, pursuing goals it can never reach or taking irreversible steps with real-world consequences. In safety-critical settings, such failures may further lead the agent to select irreversible or high-cost actions based on predicted future states that are unattainable in practice~\citep{amodei2016concrete, berkenkamp2017safe}.

This issue becomes critical when agent actions alter the future executable action sets, we term these \textbf{Structure-Changing Events (SC Events)}. Such events are common in domains with compositional Prerequisites. For example, crafting a wooden pickaxe unlocks stone mining in Craftax \citep{hafner2022crafter}; drinking a potion enables lava traversal in MiniHack \citep{samvelyan2021minihack}; picking up a potato near a sink enables \textsc{clean} in ALFWorld \citep{shridhar2021alfworld}. Over long horizons, SC events can cascade, with each event expanding the set of executable actions and inducing exponential growth in the joint affordance space as compositional depth increases. Standard world models cannot reliably track which actions are executable after each SC event, and thus accumulate compounding prediction errors over multi-step rollouts. Prior work on structured world models has explored context-dependent causal graphs~\citep{hwang2024finegrained}, but these methods capture transition dynamics within fixed causal regimes. They infer causal structure from observational data without representing the preconditions that govern when each action becomes available, and therefore still cannot prevent rule violations during imagination. Fundamentally, implicit world models struggle with compositional affordance structures because they fail to answer the question "is this action allowed now?" from "what will this action do?", and thus cannot recombine known preconditions to reason about action availability at depths unseen during training.

In this paper, we propose the Affordance-Grounded World Model (AGWM) to address two failure modes of implicit world models: compounding multi-step imagination error and failure to generalize when novel rule combinations appear at test time. Rather than leaving affordance structure implicit, AGWM explicitly tracks at each timestep which affordances are currently active, which are newly reachable given satisfied prerequisites (the frontier), and how prerequisite relations are organized across the environment's DAG schema. By conditioning its dynamics on this explicit representation, AGWM adapts immediately to SC events instead of relying on latent state drift. Across multiple benchmarks, AGWM significantly reduces multi-step imagination error and generalizes to novel affordance configurations without oracle supervision.
\newline 
\newline 
Our contributions are:
\begin{itemize}[leftmargin=1.5em, itemsep=2pt, topsep=2pt]
  \item \textbf{(1) Formalization.} We formalize SC events and identify two key failure modes of implicit world models under affordance-structure change.
  \item \textbf{(2) AGWM.} We propose AGWM, combining an SC Classifier with a Dynamic Affordance Graph that enforces a frontier-mask constraint to track affordance change explicitly.
  \item \textbf{(3) Generalization.} We show empirically that the self-evolved affordance graph generalizes to novel rule combinations unseen during training.
\end{itemize}

\section{Related Work}
\label{sec:related}

World models learn environment dynamics to enable agents to plan in imagination and select optimal action sequences~\citep{ha2018world,hafner2019rssm}. DreamerV3~\citep{hafner2023dreamerv3} rolls out imagined trajectories in latent space, with follow-up works exploring Transformer, diffusion, and JEPA architectures. These approaches share a common formulation: actions condition the transition function, but the model cannot distinguish action legality from transition dynamics. To address this, MuZero~\citep{schrittwieser2020mastering} introduces legal action masks in MCTS planning, and ResWM~\citep{reswm2025} decomposes action-induced changes. However, a shared limitation remains: action legality is treated as a static function of the current state that does not evolve as the agent acts. We propose to explicitly track both the legality of current actions and the evolution of capabilities following execution, making the world model's prediction process fully traceable. 

More fundamentally, affordance modeling in RL~\citep{khetarpal2020whatcanido} formalizes valid actions conditioned on agent capabilities and improves planning efficiency, but treats affordances as static attributes not incorporated into world model dynamics. Recent visual affordance reasoning~\citep{wang2026affordancer1} similarly decouples affordances from dynamics prediction. A related line of work introduces explicit relational structure into world models via graph networks~\citep{kipf2020contrastive, huang2022cdl, li2020causal}, but the graph topology is fixed at training time and cannot evolve with environment state. Recent causal graph work recognizes that causal structure can vary with state~\citep{hwang2024finegrained, zhao2024causal}, yet switches are indexed by latent meta-states rather than causally bound to specific agent actions. 

Evolving environment dynamics have also been studied from the perspective of non-stationarity. Hidden-parameter MDPs~\citep{doshi2016hidden} and meta-RL~\citep{wang2016learning} model task variation as latent parameters or task distributions, allowing agents to adapt to different dynamics regimes. WALL-E~\citep{zhou2025walle} uses large language models to extract if-then symbolic rules from trajectories, enabling interpretable policy learning. \citet{gospodinov2024adaptive} embed non-stationarity handling directly into a DreamerV3-style world model, continuously estimating distributional drift in latent space and adapting dynamics parameters online. Each of these approaches models dynamics that shift across tasks or over time, but none treats action-triggered affordance expansion as a first-class modeling target.

The unifying blind spot across all three threads is attribution: which agent action caused this affordance change, and when? In a cascading tech tree, one action at step $t$ can activate several downstream affordance edges spanning multiple tiers; each activation is a zero-to-one transition triggered by an identifiable action, not a background dynamics drift or a latent regime shift. A frozen-graph model predicts with the topology from training time and misses newly activated edges entirely. A causal switcher represents the transition as a move between pre-fitted regime distributions, attributing the change to a latent meta-state rather than to the action. A non-stationarity model detects that dynamics changed but cannot identify the cause.
Our proposed design targets attribution directly: $g_t$ is derived per-timestep from the current observation via a fixed DAG schema, so each transition in node states and frontier mask is tied to the SC event that caused it. Because the graph explicitly encodes which affordances are currently reachable (frontier mask) and which prerequisites are satisfied (edge states), the model can represent affordance states that reflect novel combinations of rules, a capability that latent-indexed methods, bounded by their training-distribution regimes, cannot provide by construction.

\section{Method: Affordance-Grounded World Models}
\label{sec:method}

\subsection{Problem Formulation}
\label{sec:problem}

\paragraph{Structure-Changing Events.}

We consider environments modeled as a Markov decision process (MDP) $(\mathcal{S}, \mathcal{A}, T, R, \gamma)$, where $\mathcal{S}$ is the state space, $\mathcal{A}$ the action space, $T: \mathcal{S} \times \mathcal{A} \to \mathcal{S}$ the transition function, $R$ the reward function, and $\gamma \in [0,1)$ the discount factor. An action $a_t$ at state $s_t$ is \textbf{structure-changing} (SC) if it alters the affordance set:
\begin{equation}
  \text{SC}(s_t, a_t) = \mathbb{1}\left[\mathcal{F}(s_{t+1}) \neq \mathcal{F}(s_t)\right],
\end{equation}
where $\mathcal{F}(s) \subseteq \mathcal{A}$ denotes the \emph{applicable action set} at state $s$, i.e., the actions whose environmental preconditions hold. SC events are distinct from ordinary state changes: an ordinary action (e.g., moving or looking) updates the state but preserves the affordance set, while a structure-changing action (e.g., crafting or equipping) alters both. Implicit world models fail in this setting through two mechanisms. First, multi-step imagination suffers from compounding error: without an explicit affordance state, each rollout step is conditioned on stale action preconditions, and prediction error accumulates. We observe this directly on MiniHack: Vanilla's error multiplies by $73.7\times$ from step 1 to step 8, while \ours{} remains at $5.5\times$ (Table~\ref{tab:multistep}). Second, compositional generalization fails: Vanilla achieves $0\%$ CDA on KeyDungeon and $2.5\%$ on Craftax, near the random baseline of $5.9\%$, when evaluated on SC-critical decisions (Table~\ref{tab:main_results}), and collapses on out-of-distribution affordance configurations where \ours{} reaches $90.9$--$100\%$ CDA (Table~\ref{tab:ood}).

\subsection{AGWM}
\paragraph{Dynamic Affordance Graph.}

\begin{wrapfigure}[26]{r}{0.48\textwidth}
  \vspace{-12pt}
  \centering
  \includegraphics[width=1\linewidth]{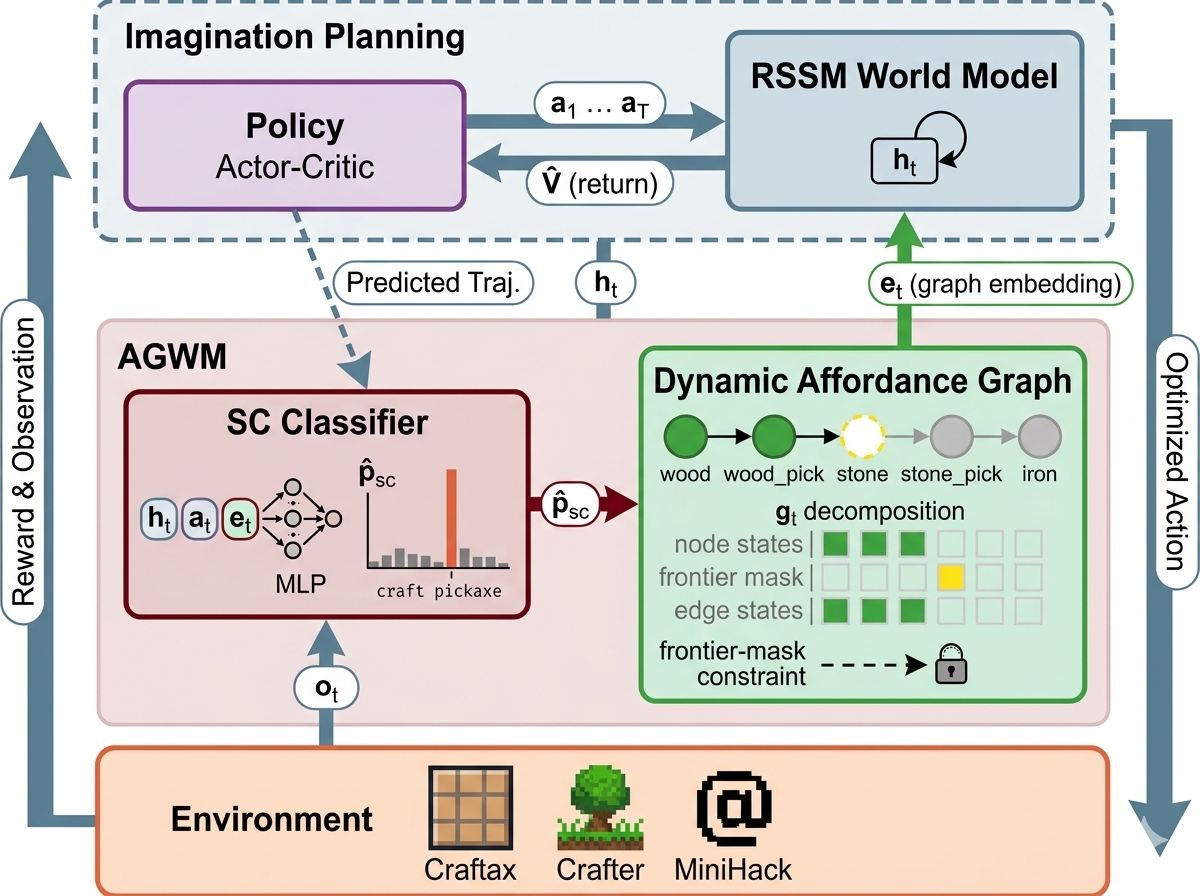}
  \caption{\textbf{AGWM system overview.} The environment delivers reward and observation to AGWM. The SC Classifier predicts whether $(h_t, a_t, e_t)$ triggers a structure-changing event and signals the Dynamic Affordance Graph to self-evolve $g_t$. The graph embedding $e_t$ conditions the RSSM World Model, gating imagination rollouts to the current affordance frontier. The Imagination Planning loop uses the imagined trajectories to optimize the Actor-Critic policy.}
  \label{fig:agwm_system}
\end{wrapfigure}

To address these problems, we represent $\mathcal{F}(s)$ as a structured feature vector $g_t \in \{0,1\}^d$, derived per-timestep from the current observation, which we call the \textbf{affordance graph}. Each environment has a fixed DAG schema defining $N$ affordance nodes (items, tools, structures) and $E$ prerequisite edges (tech-tree dependencies); $g_t$ concatenates three binary components: \emph{node states} ($N$ dims) encoding which affordances are currently achieved; a \emph{frontier mask} ($N$ dims) encoding which affordances are newly reachable (all prerequisites satisfied but not yet achieved); and \emph{edge states} ($E$ dims) indicating which prerequisite edges are currently satisfied. The model learns to predict how the graph transitions at each step:
\begin{equation}
  \hat{g}_{t+1} = f_\text{graph}(h_t, a_t, g_t),
\end{equation}
where $h_t$ is the recurrent hidden state and $f_\text{graph}$ is a learned predictor. The key design property is the \emph{frontier-mask constraint}: an affordance can become active only when its DAG prerequisites are already met, which matches the OR-monotone structure of tech trees and enables $d$ independent binary predictions rather than joint modeling of the full $2^d$ affordance space.

\begin{figure}[t]
  \centering
  \includegraphics[width=\linewidth]{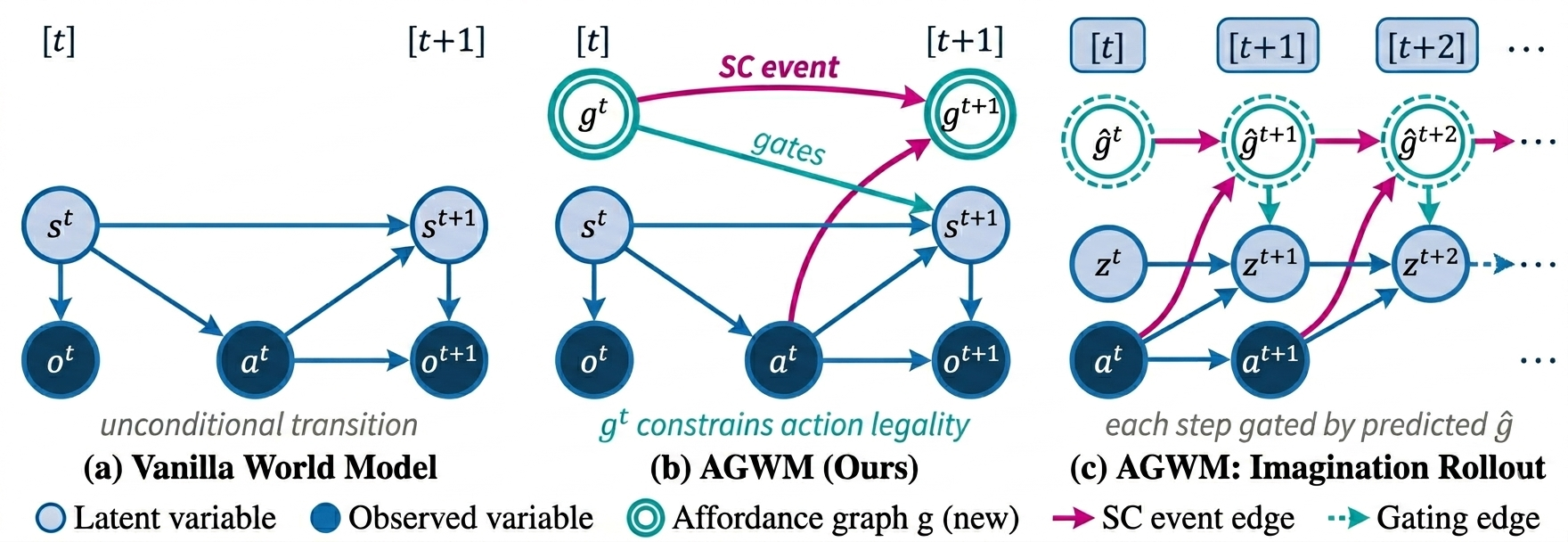}
  \caption{\textbf{Probabilistic graphical models of world model variants.}
  \textbf{(a) Vanilla world model}: $a^t$ feeds unconditionally into $s^{t+1}$; the model cannot enforce whether $a^t$ is currently executable, causing compounding imagination error after SC events.
  \textbf{(b) \ours{} (one step)}: $g^t$ is introduced as an explicit affordance variable. A structure-changing action $a^t$ triggers an SC event edge (magenta) that updates $g^{t+1}$, while $g^t$ gates the transition to $s^{t+1}$ via a gating edge (teal), enforcing action legality.
  \textbf{(c) \ours{} imagination rollout}: the predicted graph $\hat{g}^t$ (dashed) is propagated forward at each step, gating every imagined transition so that multi-step rollouts remain within the current affordance frontier.}
  \label{fig:graphical_model}
\end{figure}

Figure~\ref{fig:graphical_model} contrasts the graphical models of a conventional world model and \ours{}. In a standard POMDP world model, the action $a^t$ feeds unconditionally into the next state $s^{t+1}$: the model associates action-outcome pairs statistically but cannot enforce whether $a^t$ is currently executable. \ours{} introduces the affordance graph $g^t$ as an explicit variable. $g^t$ gates the transition $a^t \to s^{t+1}$, preventing the model from predicting outcomes for infeasible actions. Crucially, $g^t$ is not observed but self-evolved: it is predicted from $(h_t, a_t)$ and updated monotonically as the agent discovers new affordances.

Specifically, \ours{} augments a GRU-based recurrent world model with three modules (Figure~\ref{fig:agwm_system}). The base architecture encodes observations into latent states $z_t = \text{Enc}(o_t)$, updates a recurrent hidden state $h_t = \text{GRU}([z_t, a_t], h_{t-1})$, predicts the next latent $\hat{z}_{t+1} = f_\text{dyn}(h_t)$, and reconstructs $\hat{o}_{t+1} = \text{Dec}(\hat{z}_{t+1})$.

\textbf{Graph Encoder.} The affordance graph $g_t \in \{0,1\}^d$ is embedded via a linear projection $e_t = \text{GraphEnc}(g_t) \in \mathbb{R}^{d_e}$, where $d_e = 64$ across all environments. This embedding is injected into two places: (1) the GRU input, extending it to $h_t = \text{GRU}([z_t, a_t, e_t], h_{t-1})$, so the recurrent state is informed by the current affordance structure; and (2) the decoder, giving $\hat{o}_{t+1} = \text{Dec}([\hat{z}_{t+1}, e_{t+1}])$, so the reconstruction loss directly backpropagates through the graph embedding. The decoder path is critical: without it, the graph enters only through the GRU, where the model can learn to ignore it. Conditioning the decoder forces the reconstruction loss to depend on graph quality, providing a strong gradient signal to the graph predictor.

\textbf{SC Classifier.} A two-layer MLP $f_\text{sc}$ takes the hidden state and action embedding as input and predicts whether $a_t$ triggers a structure change: $\hat{p}_\text{sc} = \sigma(f_\text{sc}(h_t, a_t, e_t))$. SC events are rare in practice (typically 5--15\% of steps), so we apply positive-class weighting ($w_\text{pos} = 5$) to the binary cross-entropy loss to prevent the classifier from collapsing to always-negative predictions.

\textbf{Graph Predictor.} A separate MLP $f_\text{graph}$ predicts the next affordance state per dimension: $\hat{g}_{t+1} = \sigma(f_\text{graph}(h_t, a_t, e_t)) \in [0,1]^d$. Each dimension is predicted independently, decomposing the $2^d$ joint affordance space into $d$ binary problems. At inference, the predicted logits are thresholded and combined with the current graph via the monotonicity constraint (Section~\ref{sec:self_evolving}). Because $e_t = \text{GraphEnc}(g_t)$ enters the predictor input, a gradient shortcut through $e_t$ is possible in principle; however, the 10.0\% Aff.\ Acc improvement over Vanilla on Craftax (Table~\ref{tab:main_results}) confirms that the predictor learns meaningful affordance dynamics rather than trivially copying its input.

\begin{figure}[H]
  \centering
  \includegraphics[width=0.95\linewidth]{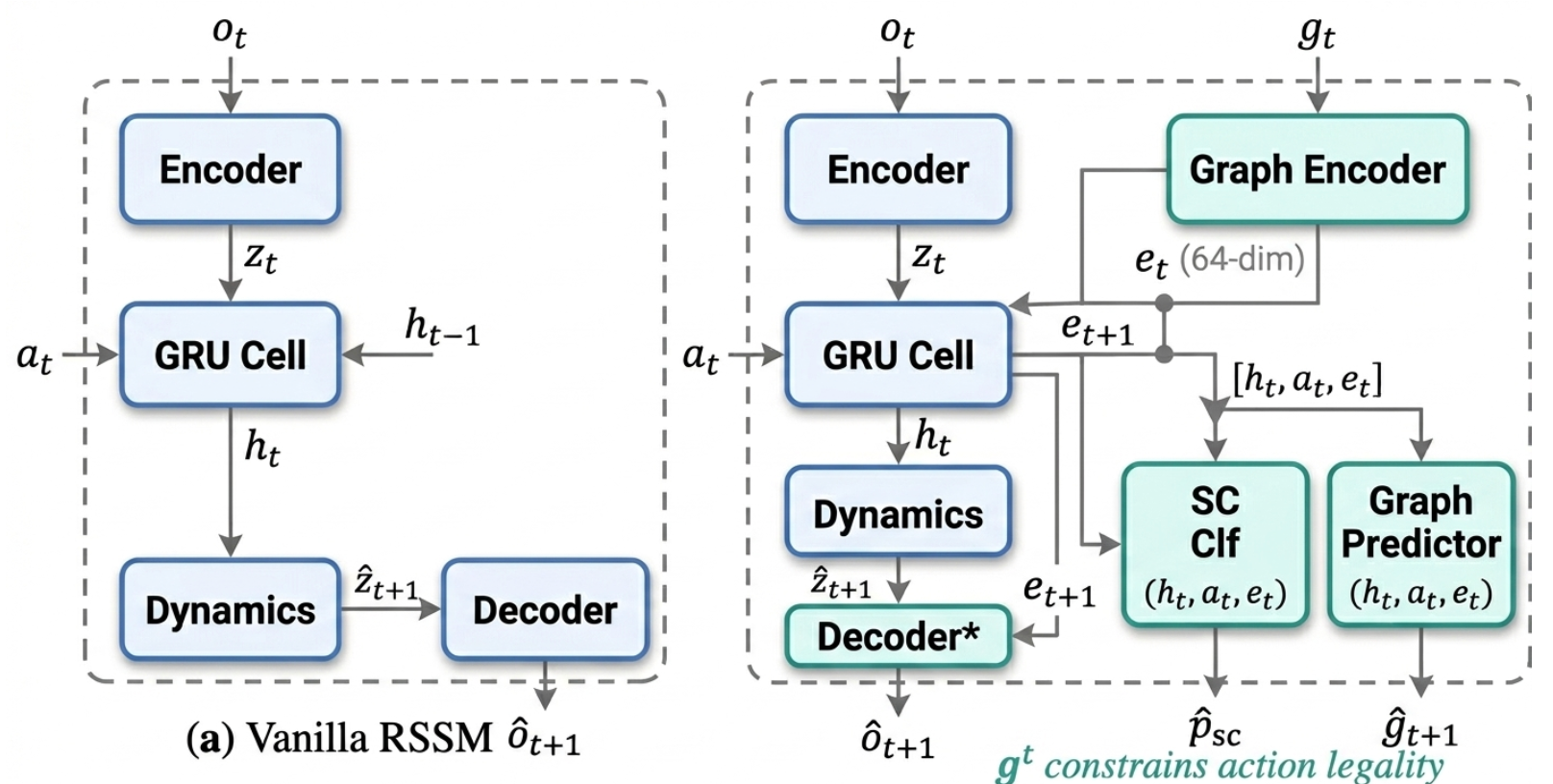}
  \caption{\textbf{Architecture comparison.} (a) Vanilla RSSM processes observations and actions through a GRU. (b) \ours{} augments the RSSM with a self-evolving affordance graph: the Graph Encoder embeds affordance structure into the GRU input and decoder, while the SC Classifier and Graph Predictor auxiliary heads learn to detect and predict structure changes.}
  \label{fig:architecture}
\end{figure}
\vspace{-0.6cm}

\subsection{Self-Evolving Affordance Discovery}
\label{sec:self_evolving}

Unlike prior affordance models that assume a fixed or oracle-provided mapping, \ours{}'s affordance graph is \emph{self-evolved} from agent experience, oracle-free at test time.

\textbf{SC label generation.} At each training step, the SC label is computed by comparing consecutive affordance states: $p_\text{sc} = \mathbf{1}[g_{t+1} \neq g_t]$. The per-dimension graph target is $g_{t+1}$ itself, obtained by comparing $\mathcal{F}(s_t)$ and $\mathcal{F}(s_{t+1})$ directly from environment state. This requires access to the affordance function $\mathcal{F}$ during training but not at test time, where $g_t$ is maintained by the model's own predictions.

\begin{wrapfigure}{r}{0.55\textwidth}
  \vspace{-8pt}
  \centering
  \includegraphics[width=\linewidth]{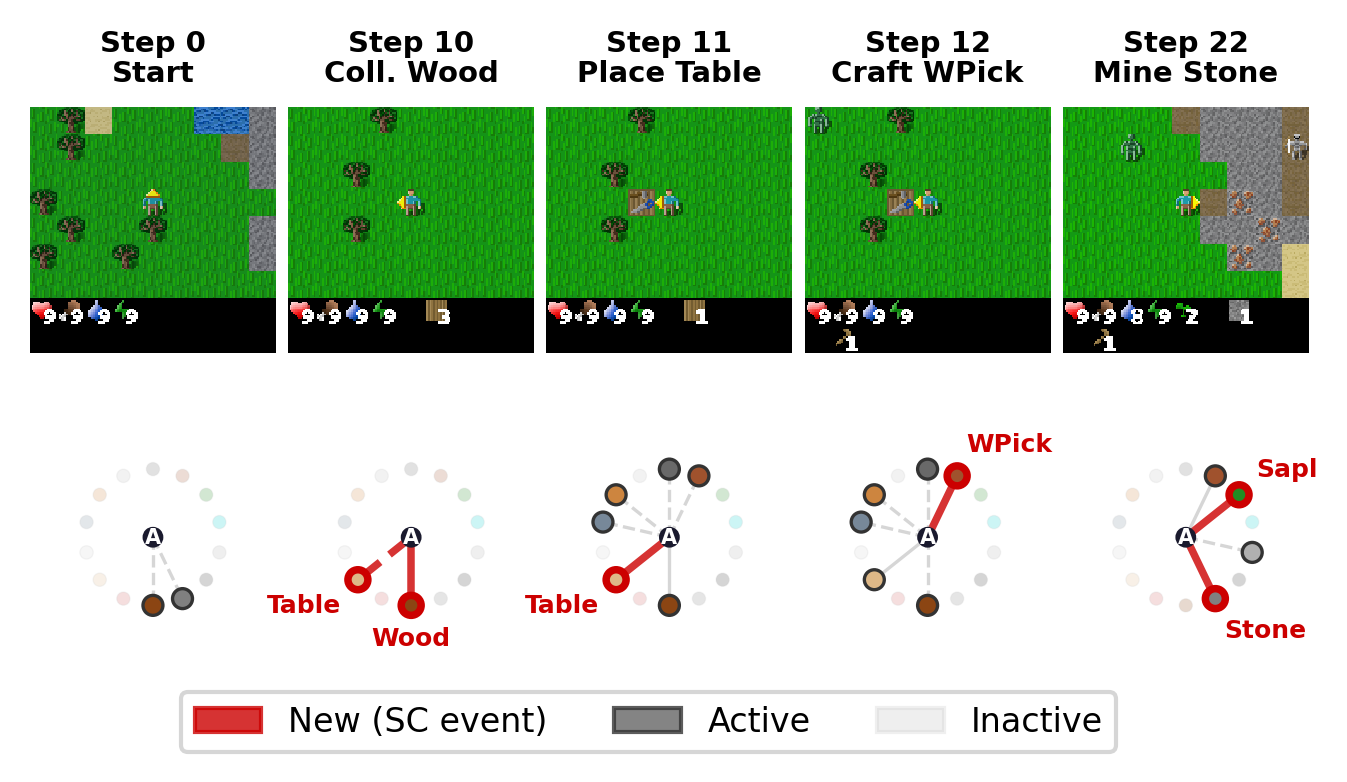}
  \caption{\textbf{Affordance graph evolution in Craftax.} As the agent progresses through the tech tree within an episode, the node-state and frontier-mask components of $g_t$ update to reflect newly achieved affordances and currently reachable next steps; the graph predictor learns to anticipate these transitions from $(h_t, a_t, g_t)$.}
  \label{fig:graph_evolution}
  \vspace{-6pt}
\end{wrapfigure}
\textbf{Frontier-mask constraint.} Affordances in tech-tree environments follow prerequisite ordering: stone is mineable only after a wood pickaxe is crafted. The frontier mask encodes this structurally: the frontier bit for node $v$ is 1 if and only if all parent prerequisites of $v$ are satisfied in the current node-state vector. The graph predictor therefore only needs to activate nodes whose prerequisites are already met, restricting the prediction space to the DAG's reachable frontier at each step. This constraint is computed analytically from the fixed DAG schema and requires no additional learning, preventing the model from producing physically impossible affordance combinations under distribution shift.

\textbf{Graph evolution within an episode.} Because $g_t$ is computed from the current observation at each step, it naturally tracks the agent's affordance state as it progresses through the tech tree within an episode. As the agent crafts tools and gathers resources, the frontier mask expands to reflect newly reachable affordances. Figure~\ref{fig:graph_evolution} illustrates this evolution alongside the agent's progression in Craftax.

\subsection{Training Objective}
\label{sec:training}

The total loss combines reconstruction, dynamics, SC classification, and graph prediction:
\begin{equation}
  \mathcal{L} = \mathcal{L}_\text{recon} + \mathcal{L}_\text{dyn} + \lambda_\text{SC}\,\mathcal{L}_\text{SC} + \lambda_\text{graph}\,\mathcal{L}_\text{graph},
\end{equation}
where each term is:
\begin{alignat}{2}
  \mathcal{L}_\text{recon} &= \text{MSE}(\hat{o}_{t+1}, o_{t+1}), &\qquad
  \mathcal{L}_\text{dyn}   &= \text{MSE}(\hat{z}_{t+1}, z_{t+1}),  \\
  \mathcal{L}_\text{SC}    &= \text{BCE}(\hat{p}_\text{sc}, p_\text{sc}), &\qquad
  \mathcal{L}_\text{graph} &= \tfrac{1}{d}\textstyle\sum_{i=1}^{d} \text{BCE}(\hat{g}_{t+1}^{(i)}, g_{t+1}^{(i)}).
\end{alignat}

The graph loss averages over all $d$ dimensions independently. We set $\lambda_\text{SC} = 1.0$ and $\lambda_\text{graph} = 2.0$. Results are robust to both hyperparameters: degradation is stable across $\lambda_\text{graph} \in [0.5, 5.0]$ and across graph embedding dimensions $d_e \in \{8, 16, 32, 64\}$, with $\lambda_\text{graph} = 0$ the only configuration that causes substantial degradation (see Appendix~\ref{app:hyperparam}). The reconstruction and dynamics losses are unweighted relative to each other.

\section{Experiments}
\label{sec:experiments}

We evaluate \ours{} by examining the following research questions:
(1) Does \ours{} reduce multi-step imagination error over the Vanilla baseline, and does the advantage grow with longer rollout horizons? (Table~\ref{tab:mse}, Table~\ref{tab:multistep})
(2) Does it accurately detect SC events and learn interpretable affordance structure? (Table~\ref{tab:main_results}, Figure~\ref{fig:graph_evolution})
(3) Does the self-evolved graph generalize to novel rule combinations not seen during training? (Table~\ref{tab:ood})
(4) Are SC detection and graph conditioning both necessary? (Table~\ref{tab:graph_ablation})

\subsection{Benchmarks}
\label{sec:benchmarks}

Full environment descriptions are provided in Appendix~\ref{app:envs}. Briefly: \textbf{KeyDungeon}, \textbf{Forage}, and \textbf{Harvest} are custom pixel gridworlds at $D{=}1$ where SC events erase colored squares from the observation. \textbf{Relay} ($D{=}3$) and \textbf{Cascade} ($D{=}4$) extend this with chain-unlocking mechanics that produce two simultaneous pixel changes per SC event. \textbf{MiniHack}~\citep{samvelyan2021minihack} (LavaCross, $D{=}2$), \textbf{Crafter}~\citep{hafner2022crafter} ($D{=}3$), and \textbf{Craftax}~\citep{matthews2024craftax} ($D{=}4$) are standard benchmarks included for breadth and comparability with prior work.

\subsection{Implementation Details}

All three variants share the RSSM backbone from DreamerV3~\citep{hafner2023dreamerv3,hafner2019rssm} and identical capacity (within 17\%). \textbf{Vanilla} is the unmodified RSSM. \sconly{} adds the SC Classifier head to Vanilla (Section~\ref{sec:method}) but \emph{without} the affordance graph; it isolates whether SC event detection alone (without explicit graph conditioning) suffices. \textbf{\ours{}} is the full model: SC Classifier plus Graph Encoder, Graph Predictor, and frontier-mask constraint. All models are trained with AdamW (lr=$5\times10^{-4}$, weight decay=$10^{-4}$), gradient clipping (norm 1.0), on 16-step sequences with batch size 64, for 500k steps. The observation encoder and decoder are environment-specific: CNN for pixel observations (all environments except Craftax) and MLP for symbolic observations (Craftax). All results report 3-seed mean $\pm$ std.

Prediction quality is measured via imagination MSE (8-step rollout after 10-step posterior warmup), counterfactual detection accuracy (CDA), counterfactual prediction gap, and affordance prediction accuracy. For Vanilla, CDA is measured by training a linear probe on frozen hidden states; for \ours{} and \sconly{}, CDA uses the trained SC Classifier directly.

\subsection{World Model Quality}
\label{sec:results_wm}

\begin{table}[ht]
  \caption{Imagination MSE ($\times 10^{-3}$, lower is better). \textbf{Avg}: step-averaged over the full rollout; \textbf{Step~1}/\textbf{Step~8}: short- and long-horizon error accumulation. \ours{} uses a self-evolved graph (no oracle). \underline{\textbf{Bold+underline}}: best per column; \textbf{bold}: second best. 3-seed mean $\pm$ std.}
  \label{tab:mse}
  \label{tab:multistep}
  \centering
  \resizebox{\linewidth}{!}{%
  \begin{tabular}{llcccccc}
    \toprule
    & & \multicolumn{3}{c}{Vanilla} & \multicolumn{3}{c}{\ours{} (self-evolved)} \\
    \cmidrule(lr){3-5} \cmidrule(lr){6-8}
    Environment & Depth & Avg & Step 1 & Step 8 & Avg & Step 1 & Step 8 \\
    \midrule
    KeyDungeon           & Low     & $10.64\pm0.80$                              & $8.53\pm1.90$                               & $12.39\pm0.35$                              & $\underline{\mathbf{5.63\pm1.44}}$    & $\underline{\mathbf{2.35\pm0.72}}$    & $\underline{\mathbf{9.05\pm1.80}}$    \\
    Forage               & Low     & $11.58$                                     & $20.11$                                     & $4.33$                                      & $\underline{\mathbf{6.95}}$           & $\underline{\mathbf{12.47}}$          & $\underline{\mathbf{2.27}}$           \\
    Harvest              & Low     & $263.4$                                     & $313.9$                                     & $218.5$                                     & $\underline{\mathbf{21.19}}$          & $\underline{\mathbf{36.90}}$          & $\underline{\mathbf{7.72}}$           \\
    MiniHack$^\dagger$   & Low-Med & $1496.2\pm1116.5$                           & $\underline{\mathbf{238.2}}$                & $17564.8$                                   & $\underline{\mathbf{1021.2\pm102.4}}$ & $388.0$                               & $\underline{\mathbf{2148.4}}$         \\
    Relay                & Medium  & $621.3$                                     & $615.8$                                     & $627.1$                                     & $\underline{\mathbf{6.11}}$           & $\underline{\mathbf{9.74}}$           & $\underline{\mathbf{3.17}}$           \\
    Crafter              & Medium  & $\underline{\mathbf{5.07\pm0.19}}$          & $\underline{\mathbf{3.45\pm0.06}}$          & $\underline{\mathbf{7.27\pm0.57}}$          & $5.56\pm0.03$                         & $3.83\pm0.07$                         & $8.28\pm0.19$                         \\
    Cascade              & High    & $205.6$                                     & $305.2$                                     & $116.4$                                     & $\underline{\mathbf{16.78}}$          & $\underline{\mathbf{26.95}}$          & $\underline{\mathbf{7.67}}$           \\
    Craftax$^\ddagger$   & High    & $\underline{\mathbf{0.0041\pm0.0012}}$      & $\underline{\mathbf{0.0040\pm0.0012}}$      & $\underline{\mathbf{0.0041\pm0.0012}}$      & $0.0894\pm0.0049$                     & $0.0875\pm0.0045$                     & $0.0919\pm0.0052$                     \\
    \bottomrule
  \end{tabular}}
  \par\smallskip
  {\small $^\dagger$MiniHack: glyph-embedding space (16-dim $\times$ 21$\times$79 grid), not comparable to pixel/symbolic rows.
  $^\ddagger$Craftax: 256-dim obs-embedding space; DR $\approx$ 1.0 for both models (no horizon degradation); CDA is the primary differentiator (Table~\ref{tab:main_results}).}
\end{table}

Graph conditioning helps most when SC events cause large, visible changes to the observation. Our $2{\times}2$ design isolates this: Relay and Cascade have the same compositional depth as Crafter and Craftax, but with clear pixel changes at every SC event, \ours{} achieves $101.7\times$ and $12.3\times$ lower MSE. At the same depth, Crafter and Craftax show no improvement because their SC signals are subtle or symbolic. When crafting an item changes only a small inventory icon, even a model with the correct affordance graph cannot improve its pixel predictions enough to reduce overall MSE. The difference grows over longer rollouts: on MiniHack, Vanilla reaches $73.7\times$ of its step-1 error by step 8, while \ours{} stays at $5.5\times$. This happens because errors build up: without knowing what the new affordance state is, each step starts from a wrong assumption about which actions are available.

\paragraph{SC detection and affordance understanding.}

\begin{table}[ht]
  \caption{SC Detection Accuracy (CDA, \%, higher is better), probing results, and OOD generalization (3-seed mean $\pm$ std). Aff.\ Acc: next-step affordance prediction accuracy (\%); CF Gap: L2 distance between SC and non-SC predictions. $^\ddagger$OOD splits (\ours{} only, Vanilla/+SC not evaluated on held-out configs): KD~L4 = Levels 1--3 $\to$ L4 cross-color; KD~L5 = Levels 1--4 $\to$ L5 chain4; Craftax = Tech tiers 0--1 $\to$ tiers 2--3.}
  \label{tab:main_results}
  \label{tab:ood}
  \centering
  \footnotesize
  \setlength{\tabcolsep}{4pt}
  \begin{tabular}{llccc}
    \toprule
    Metric & Environment & Vanilla              & \sconly{}                           & \ours{}                       \\
    \midrule
    \multirow{4}{*}{CDA (\%) $\uparrow$}
           & KeyDungeon  & $0.0$                & $\underline{\mathbf{100.0}}$  & $\underline{\mathbf{100.0}}$  \\
           & Crafter     & $0.0\pm0.0$          & $\underline{\mathbf{24.2\pm22.7}}$ & $0.0\pm0.0$               \\
           & MiniHack    & $0.0$                & $0.0$                         & $0.0$                         \\
           & Craftax     & $2.5\pm0.2$          & $\underline{\mathbf{99.6\pm0.2}}$ & $\mathbf{98.6\pm0.8}$      \\
    \midrule
    \multirow{3}{*}{Aff.\ Acc (\%) $\uparrow$}
           & KeyDungeon  & $87.4\pm0.0$         & $90.9\pm0.3$                  & $\underline{\mathbf{99.9\pm0.2}}$  \\
           & Crafter     & $\underline{\mathbf{100.0\pm0.0}}$ & $\underline{\mathbf{100.0\pm0.0}}$ & $\underline{\mathbf{100.0\pm0.0}}$ \\
           & Craftax     & $90.8\pm0.0$         & $89.7\pm0.1$                  & $\underline{\mathbf{99.6\pm0.0}}$  \\
    \midrule
    \multirow{2}{*}{CF Gap $\uparrow$}
           & KeyDungeon  & $0.08\pm0.01$        & $0.80\pm0.15$                 & $\underline{\mathbf{1.64\pm0.20}}$ \\
           & Craftax     & $0.00\pm0.00$        & $0.01\pm0.00$                 & $\underline{\mathbf{0.02\pm0.00}}$ \\
    \midrule
    \multirow{3}{*}{OOD CDA (\%) $\uparrow$}
           & KD L4$^\ddagger$   & -- & -- & $\underline{\mathbf{90.9}}$ \\
           & KD L5$^\ddagger$   & -- & -- & $\underline{\mathbf{100.0}}$ \\
           & Craftax$^\ddagger$ & -- & -- & $\underline{\mathbf{87.9\pm0.7}}$ \\
    \bottomrule
  \end{tabular}
\end{table}

As the affordance structure gets deeper, a fixed-size latent vector finds it harder to keep track of it. On Craftax, where four prerequisite tiers must be maintained at the same time, Vanilla scores only $2.5\%$ CDA while \ours{} reaches $98.6\%$; a hidden state just does not have enough room to store that much structured information. On Crafter, \sconly{} reaches $24.2\%$ CDA (vs.\ $0\%$ for \ours{}) but with high variance ($\pm22.7\%$, driven by a single seed at $54.5\%$) and no improvement in MSE, suggesting the SC head sometimes detects events that are not real affordance transitions. The full model avoids this because any detected SC event must update the graph, and the model is then judged on whether those updates lead to better predictions. The graph also makes the model's internal state readable: on KeyDungeon, \ours{}'s predicted observations differ $20.5\times$ more between SC and non-SC actions than Vanilla's ($1.64$ vs.\ $0.08$ CF Gap). Figure~\ref{fig:graph_evolution} shows each unlock as a single added edge, with no probing required to see what the model knows.

\begin{table}[ht]
  \centering
  \caption{Graph source ablation on KeyDungeon. All \ours{} variants share the same trained model. MSE ($\times10^{-3}$), mean $\pm$ std, 3 seeds.}
  \label{tab:graph_ablation}
  \small
  \begin{tabular}{lc}
    \toprule
    Graph source            & MSE ($\times10^{-3}$)                \\
    \midrule
    None (Vanilla)          & $9.19 \pm 0.23$                      \\
    Zeroed $g$ (\ours{})    & $19.54 \pm 2.06$                     \\
    Frozen $g$ (\ours{})    & $5.76 \pm 0.62$                      \\
    Predicted $g$ (\ours{}) & $\mathbf{5.23 \pm 0.51}$             \\
    Oracle $g$ (\ours{})    & $\underline{\mathbf{4.95 \pm 0.61}}$ \\
    \bottomrule
  \end{tabular}
\end{table}

\paragraph{Compositional generalization.} The learned graph works well on affordance rules not seen during training. On KeyDungeon, \ours{} reaches $90.9\%$ CDA on cross-color rules and $100.0\%$ on entirely new L5 chain mechanisms (Table~\ref{tab:ood}); on Craftax, training on tiers 0--1 and testing on tiers 2--3 gives $87.9\pm0.7\%$ OOD CDA. The $100\%$ accuracy on new L5 chains suggests the graph has learned general affordance rules (``collecting \emph{some} item enables \emph{some} next item'') rather than memorizing specific item pairs, so it handles new item combinations correctly. The oracle control experiment (Appendix~\ref{app:oracle_control}) shows why this works: the self-evolved graph ($1.34\times$ degradation) is close to the oracle ceiling ($1.12\times$) without any ground-truth graph at test time, while zeroing the graph ($6.06\times$) performs worst, confirming that the self-evolving mechanism is the main source of test-time accuracy.

\paragraph{Ablations.} Table~\ref{tab:graph_ablation} shows that removing either component hurts. Without graph-conditioned decoding, the graph becomes a passive input the model can learn to ignore, which is why \sconly{} outperforms the full \ours{} in that setting. Adding the decoding loss brings the benefit back. Even a frozen graph (fixed at its initial state, $5.76$) already beats Vanilla ($9.19$), showing that any reasonable affordance structure is more useful than none. The self-evolving mechanism then closes most of the gap to the oracle ($5.23$ vs.\ $4.95$). Giving the model an all-zero graph at test time ($19.54$) is worse than no graph at all ($9.19$), since the model was never trained on that input. Any non-zero $\lambda$ helps, and the gain levels off above $0.5$.

\paragraph{Sample efficiency.} With less training data, the benefit of explicit SC structure gets larger. At 25\% data, \ours{} and \sconly{} both beat Vanilla ($14.94$ and $14.76$ vs.\ $16.64$); at full data all three models perform about the same (Table~\ref{tab:sample_efficiency}). Graph accuracy also levels off early: Aff.Acc reaches $97.2\%\pm1.8\%$ at just 25\% data and goes above $98\%$ at 50\%, showing that the affordance transition structure can be learned from a small number of trajectories.

\begin{table}[ht]
  \caption{Sample efficiency on KeyDungeon. Imagination MSE ($\times10^{-3}$) and affordance-graph accuracy (Aff.Acc) vs.\ training data fraction. Mean $\pm$ std, 3 seeds. MSE lower is better; Aff.Acc higher is better.}
  \label{tab:sample_efficiency}
  \centering
  \small
  \begin{tabular}{lccccc}
    \toprule
    Data  & $n_{\text{seq}}$ & Vanilla MSE             & \sconly{} MSE                         & \ours{} MSE                           & \ours{} Aff.Acc    \\
    \midrule
    25\%  & 28  & $16.64 \pm 1.76$          & $\underline{\mathbf{14.76 \pm 0.57}}$ & $\mathbf{14.94 \pm 0.40}$             & $97.2\% \pm 1.8\%$ \\
    50\%  & 56  & $13.01 \pm 0.08$          & $13.31 \pm 0.38$                      & $\underline{\mathbf{12.78 \pm 0.47}}$ & $98.5\% \pm 0.1\%$ \\
    75\%  & 84  & $12.28 \pm 0.12$          & $12.38 \pm 0.23$                      & $\underline{\mathbf{12.12 \pm 0.14}}$ & $99.4\% \pm 0.7\%$ \\
    100\% & 112 & $\mathbf{12.26 \pm 0.13}$ & $\underline{\mathbf{12.24 \pm 0.32}}$ & $12.41 \pm 0.31$                      & $99.5\% \pm 0.7\%$ \\
    \bottomrule
  \end{tabular}
\end{table}

\vspace{0.3cm}
\section{Conclusion}
\label{sec:conclusion}

We introduced \ours{}, a world model that explicitly tracks how agent actions change environment affordances through a self-evolving affordance graph.
Across our benchmark suite, we demonstrated that explicit SC detection and graph-conditioned imagination reduce multi-step prediction error; that the self-evolved graph generalizes to novel rule combinations without oracle supervision; and that the frontier-mask constraint is essential: removing it causes performance worse than the graph-free Vanilla baseline due to distribution shift at test time.

\section*{Limitations.} The graph dimensions (which affordance types to track) are currently hand-designed per environment. Learning the graph schema from data is an important direction. The self-evolving mechanism also requires comparing consecutive states to detect SC events; extending to delayed or partial observability is future work.

\section*{Broader impact.} This work proposes a method for learning world models that explicitly track which agent actions remain executable as the environment's rule structure evolves. The primary intended application is model-based reinforcement learning in compositional domains such as open-ended game environments and robotic task planning. On the positive side, better world models that respect action preconditions could improve the safety and reliability of autonomous agents deployed in structured environments, since the agent's imagination is constrained to feasible action sequences rather than arbitrary counterfactuals. On the negative side, more capable planning agents could be applied to adversarial or harmful tasks. We note that the current work is evaluated exclusively on simulated benchmark environments; deployment in physical systems would require additional safety validation beyond what this paper addresses. We do not anticipate near-term negative societal impacts specific to this work.

\appendix

\section{Implementation Details}
\label{app:impl}

\paragraph{Network architecture.}
All models share an identical base architecture.
The observation encoder is a 3-layer CNN (channels 32/64/64, kernel 3, stride 2) for pixel observations (Crafter, KeyDungeon), a glyph-embedding CNN (16-dim per glyph, then 3 Conv layers) for MiniHack, and a 2-layer MLP (hidden dim 256) for symbolic observations (Craftax). The GRU has hidden size 512. The dynamics network $f_\text{dyn}$ and decoder are symmetric 2-layer MLPs with hidden dim 512. For \ours{}, the Graph Encoder is a single linear layer mapping $g_t \in \{0,1\}^d$ to $e_t \in \mathbb{R}^{64}$. The SC Classifier and Graph Predictor are each 2-layer MLPs (hidden dim 256, ReLU) taking $[h_t, a_t, e_t]$ as input.
Total parameter counts on KeyDungeon: Vanilla 1.76M, \sconly{} 1.77M, \ours{} 1.89M; on Crafter: Vanilla 1.60M with proportionally similar scaling. \ours{} adds at most 7\% parameters over Vanilla.

\paragraph{Training.}
All models are trained for 100 epochs on pre-collected expert trajectories (500 train / 100--200 eval per environment). We use the AdamW optimizer with learning rate $5 \times 10^{-4}$, weight decay $10^{-4}$, and gradient clipping at norm 1.0. Batch size is 16 sequence windows. Loss weights: $\lambda_\text{SC} = 1.0$, $\lambda_\text{graph} = 2.0$. The SC classifier uses positive-class weighting $w_\text{pos} = 5$ to compensate for class imbalance (SC events occur in 5--15\% of steps depending on environment).
The affordance graph $g_t$ is computed per-timestep from the current observation via the environment's fixed DAG schema (node states, frontier mask, edge states); graph targets for training are derived directly from environment state, requiring no manual annotation. Seed 42, 123, 456 are used for reproducibility; we report 3-seed mean $\pm$ std throughout.

\paragraph{Evaluation.}
Imagination MSE is measured over 8-step rollouts following a 10-step posterior warmup from ground-truth observations. All metrics are reported as mean $\pm$ std over 3 independent seeds with different random initializations.

\paragraph{Environment-specific details.}

\section{Environment Descriptions}
\label{app:envs}

\begin{table}[ht]
  \caption{Environment overview. Compositional depth $D$ = maximum cascading SC chain length. $^\dagger$This work.}
  \label{tab:envs}
  \centering
  \resizebox{\linewidth}{!}{%
  \begin{tabular}{llp{4.5cm}cc}
    \toprule
    Environment & Act. & Observation (encoding) & SC Type & $D$ \\
    \midrule
    KeyDungeon$^\dagger$                          & 7  & 48$\times$48$\times$3 RGB; CNN            & key$\to$door flip            & 1 \\
    Forage$^\dagger$                              & 4  & 64$\times$64$\times$3 RGB; CNN            & item collection ($\times$6)  & 1 \\
    Harvest$^\dagger$                             & 4  & 64$\times$64$\times$3 RGB; CNN            & item collection ($\times$8)  & 1 \\
    MiniHack~\citep{samvelyan2021minihack}        & 77 & 21$\times$79 glyph map; CNN              & item$\to$ability             & 2 \\
    Relay$^\dagger$                               & 4  & 64$\times$64$\times$3 RGB; CNN            & chain reveal                 & 3 \\
    Crafter~\citep{hafner2022crafter}             & 17 & 64$\times$64$\times$3 RGB; CNN            & tech tree (subtle)           & 3 \\
    Cascade$^\dagger$                             & 4  & 64$\times$64$\times$3 RGB; CNN            & chain reveal                 & 4 \\
    Craftax~\citep{matthews2024craftax}           & 17 & 1345-dim symbolic; MLP                   & cascading (symbolic)         & 4 \\
    \bottomrule
  \end{tabular}}
\end{table}

\paragraph{KeyDungeon (this work, $D{=}1$).}
A custom 48$\times$48 pixel gridworld where the agent collects a key to open a locked door and retrieves a chest. A single affordance transition (key held $\to$ door traversable) defines the sole SC event per episode. Item acquisition produces a discrete, visible pixel change: the key sprite disappears and the door glyph switches to an open doorway. KeyDungeon isolates the minimal SC setting, providing a controlled testbed for single-transition affordance prediction.

\paragraph{Forage (this work, $D{=}1$).}
A 64$\times$64 RGB gridworld with six colored items on an 8$\times$8 grid. Collecting an item causes its colored square to disappear (SC event). The binary affordance graph $g_t \in \{0,1\}^6$ records which items have been collected; a scripted policy with 15\% random noise yields SC events at $\approx$5\% of steps. Forage provides a controlled pixel-change setting: the decoder conditioned on predicted $g_{t+1}$ can anticipate item disappearance directly.

\paragraph{Harvest (this work, $D{=}1$).}
Extends Forage to eight items with two additional item colors, raising the SC rate to 6.7\% and increasing simultaneous-item-tracking complexity. The graph $g_t \in \{0,1\}^8$; all other mechanics identical to Forage.

\paragraph{Relay (this work, $D{=}3$).}
A 64$\times$64 RGB gridworld with a single chain of four items ($D{=}3$, three SC transitions). Collecting item $k$ desaturates it to a neutral used-marker color \emph{and} simultaneously reveals item $k{+}1$ from gray (locked) to its assigned bright color, producing two large pixel changes per SC event. The affordance graph $g_t \in \{0,1\}^4$ encodes which items have been collected (monotone: bits only transition $0\!\to\!1$). A scripted policy targeting the nearest visible item yields SC events at $\approx$6.7\% of steps.

\paragraph{Cascade (this work, $D{=}4$).}
Extends Relay to a single chain of five items ($D{=}4$, four SC transitions, $g_t \in \{0,1\}^5$). The same dual-change SC mechanism applies: collecting item $k$ desaturates it to a used marker while item $k{+}1$ becomes visible. SC rate $\approx$8.3\%. Cascade provides a pixel-observation counterpart to Craftax at matching compositional depth.

\paragraph{MiniHack~\citep{samvelyan2021minihack} ($D{=}2$).}
LavaCross task: the agent acquires a levitation potion and consumes it to cross lava. SC event = potion consumption activating the \texttt{levitation} affordance. Observations are $21\times79$ glyph maps; affordance changes manifest as inventory transitions with no dramatic pixel shift. Tests prediction when SC events are detectable but not visually salient.

\paragraph{Crafter~\citep{hafner2022crafter} ($D{=}3$).}
A 2D survival game with a three-tier tech tree. SC events include placing a crafting table (enables wood tools), placing a furnace (enables metal processing), and crafting successive tool tiers. Observations are $64\times64$ RGB images; affordance changes produce subtle rather than dramatic pixel differences. Joint affordance space spans $2^{17}$ states.

\paragraph{Craftax~\citep{matthews2024craftax} ($D{=}4$).}
A JAX-accelerated symbolic implementation of Crafter with a four-tier cascading tech tree. Observations are 1345-dimensional symbolic vectors encoding inventory, achievements, and local map state. The symbolic observation directly encodes affordance-relevant information, so horizon-degradation ratio DR$\approx$1.0 for all models; CDA is the primary differentiator at this depth.
Craftax uses the symbolic observation format (1345-dim) with the JAX-based implementation~\citep{matthews2024craftax}, enabling fast parallel rollouts. MiniHack uses the NetHack symbolic observation format: a 21$\times$79 glyph map (each cell is a categorical glyph ID in [0, 6000)) paired with a 27-dim blstats vector; a learned glyph embedding (16-dim per glyph) followed by a CNN encoder maps this to a latent state. KeyDungeon is a custom MiniGrid-based environment with procedurally generated key-door configurations; we evaluate on held-out map layouts unseen during training.

\section{Affordance Graph Design}
\label{app:graph}

We design compact affordance graphs for each environment:

\paragraph{Craftax (60 dims, DAG).} 15 tech-tree nodes (resources, tools, structures) $\times$ 2 state types (node\_active, frontier) plus 30 directed prerequisite edges = $15 + 15 + 30 = 60$ dims. The frontier vector encodes nodes whose prerequisites are satisfied; edge states are active only when both endpoints are unlocked.

\paragraph{KeyDungeon (14 dims, DAG).} 5 nodes (find\_key, pickup\_key, reach\_door, unlock\_door, reach\_goal) $\times$ 2 state types (node\_active, frontier) plus 4 directed prerequisite edges = $5 + 5 + 4 = 14$ dims. The key SC event is picking up the key (activates \texttt{pickup\_key}, gates the frontier to \texttt{unlock\_door}) and unlocking the door (activates \texttt{unlock\_door}, expands frontier to \texttt{reach\_goal}).

\paragraph{Crafter (38 dims, DAG).} 11 tech-tree nodes $\times$ 2 state types (node\_active, frontier) plus 16 directed prerequisite edges = $11 + 11 + 16 = 38$ dims. The DAG captures Crafter's 3-tier tech tree (wood $\to$ stone $\to$ iron $\to$ diamond); SC events are tier-crossing actions (placing a table, crafting a pickaxe) that activate new frontier nodes.

\paragraph{MiniHack LavaCross (15 dims, DAG).} 6 affordance nodes (find\_potion, drink\_potion, lava\_traversable, reach\_goal, find\_stairs, reach\_stairs) $\times$ 2 state types (node\_active, frontier) plus 3 directed prerequisite edges = $6 + 6 + 3 = 15$ dims. The key SC event is drinking the levitation potion, which activates the \texttt{lava\_traversable} node and expands the frontier to include \texttt{reach\_goal}.

\section{Additional Results}
\label{app:results}

\begin{table}[ht]
  \caption{Imagination MSE ($\times 10^{-3}$) across all environments. \ours{} uses self-evolved graph. Mean $\pm$ std over 3 seeds. Red: prior architecture; blue: current DAG architecture. Entries marked ``$\cdot$'' indicate results not yet collected.}
  \label{tab:app_mse_full}
  \centering
  \small
  \begin{tabular}{lccc}
    \toprule
    Environment & Vanilla                            & \sconly{}                                & \ours{} (self-evolved)                     \\
    \midrule
    KeyDungeon  & \nval{$10.64\pm0.80$}              & \nval{$9.79\pm2.04$}               & \nval{$\mathbf{5.63\pm1.44}$}              \\
    MiniHack    & \nval{$1496.2\pm1116.5$}$^\dagger$ & \nval{$1719.7\pm1004.0$}$^\dagger$ & \nval{$\mathbf{1021.2\pm102.4}$}$^\dagger$ \\
    Crafter     & \nval{$\mathbf{5.07\pm0.19}$}      & \nval{$5.20\pm0.31$}               & \nval{$5.56\pm0.03$}                       \\
    Craftax     & \nval{$\mathbf{0.0041\pm0.0012}$}  & \nval{$0.3057\pm0.0274$}           & \nval{$0.0894\pm0.0049$}                   \\
    \bottomrule
  \end{tabular}
\end{table}

\begin{table}[ht]
  \caption{Full probing results across all environments. Mean $\pm$ std over 3 seeds.}
  \label{tab:app_probing_full}
  \centering
  \small
  \begin{tabular}{llccc}
    \toprule
    Metric & Environment & Vanilla              & \sconly{}                           & \ours{}                       \\
    \midrule
    \multirow{3}{*}{SC MSE Ratio $\uparrow$}
           & KeyDungeon  & \nval{$1.00\pm0.00$} & \nval{$0.99\pm0.05$}          & \nval{$\mathbf{1.12\pm0.03}$} \\
           & Crafter     & \nval{$1.00\pm0.00$} & \nval{$1.00\pm0.00$}          & \nval{$\mathbf{1.00\pm0.00}$} \\
           & Craftax     & \nval{$0.97\pm0.01$} & \nval{$0.87\pm0.11$}          & \nval{$\mathbf{1.27\pm0.26}$} \\
    \midrule
    \multirow{3}{*}{CF Gap $\uparrow$}
           & KeyDungeon  & \nval{$0.08\pm0.01$} & \nval{$0.80\pm0.15$}          & \nval{$\mathbf{1.64\pm0.20}$} \\
           & Crafter     & \nval{$0.01\pm0.00$} & \nval{$\mathbf{0.09\pm0.06}$} & \nval{$0.07\pm0.03$}          \\
           & Craftax     & \nval{$0.00\pm0.00$} & \nval{$0.01\pm0.00$}          & \nval{$\mathbf{0.02\pm0.00}$} \\
    \midrule
    \multirow{3}{*}{Aff.\ Acc (\%) $\uparrow$}
           & KeyDungeon  & \nval{$87.4\pm0.0$}  & \nval{$90.9\pm0.3$}           & \nval{$\mathbf{99.9\pm0.2}$}  \\
           & Crafter     & \nval{$100.0\pm0.0$} & \nval{$100.0\pm0.0$}          & \nval{$\mathbf{100.0\pm0.0}$} \\
           & Craftax     & \nval{$90.8\pm0.0$}  & \nval{$89.7\pm0.1$}           & \nval{$\mathbf{99.6\pm0.0}$}  \\
    \bottomrule
  \end{tabular}
\end{table}

\section{Oracle Graph Control Experiment}
\label{app:oracle_control}

To rule out the alternative explanation that the oracle model underperforms on novel rules due to stale graph inputs (i.e., the graph was fixed at training time and does not reflect novel-rule affordances), we run a controlled ablation on KeyDungeon novel-rule test sets. We compare four oracle graph conditions at test time:

\begin{table}[ht]
  \caption{8-step degradation ratio on KeyDungeon novel rules under different oracle graph conditions. Mean $\pm$ std over 3 seeds. Lower is better.}
  \label{tab:oracle_control}
  \centering
  \small
  \begin{tabular}{lcc}
    \toprule
    Graph Condition & Degradation Ratio            & Description                        \\
    \midrule
    Oracle-Current  & \nval{$1.12 \pm 0.02\times$} & GT graph for novel rules (correct) \\
    Oracle-Stale    & \nval{$1.17 \pm 0.02\times$} & Frozen training-time graph (stale) \\
    Oracle-Zero     & \nval{$6.06 \pm 1.25\times$} & All-zero graph                     \\
    Self-Evolved    & \nval{$1.34 \pm 0.31\times$} & Model's own predicted graph        \\
    \bottomrule
  \end{tabular}
\end{table}

The ground-truth graph (Oracle-Current, $1.12\times$) achieves the lowest degradation, confirming that correct affordance conditioning reduces imagination error and establishing Oracle AGWM as an upper bound. The stale training-time graph ($1.17\times$) performs close to Oracle-Current: although semantically incorrect for novel rules, it preserves the dimension-activation patterns seen during training and introduces minimal distributional shift. The self-evolved graph ($1.34\times$) closely approaches the oracle upper bound without ground-truth access; the $0.22\times$ gap demonstrates that the self-evolving mechanism captures novel rule structure effectively. All-zero graph conditioning ($6.06\times$) is the worst condition: the model was trained with non-zero graphs, so an all-zero input at test time constitutes an OOD signal that severely disrupts imagination.

\section{Hyperparameter Sensitivity}
\label{app:hyperparam}

We evaluate sensitivity to the graph loss weight $\lambda_\text{graph}$ and the graph embedding dimension $d_e$ on Craftax ($D{=}4$), using 8-step degradation ratio and affordance prediction accuracy (Aff.\ Acc) as metrics. Results are reported as mean $\pm$ std over 3 seeds.

\begin{table}[ht]
  \caption{Effect of graph loss weight $\lambda_\text{graph}$ (graph embedding dimension fixed at $d_e = 32$). Degradation ratio = MSE at step 8 / MSE at step 1; lower is better.}
  \label{tab:lambda_sweep}
  \centering
  \small
  \begin{tabular}{ccc}
    \toprule
    $\lambda_\text{graph}$ & Degradation Ratio                     & Aff.\ Acc                       \\
    \midrule
    0.0  & $2.53 \pm 0.20\times$          & $0.460$ \\
    0.1  & $2.00 \pm 0.15\times$          & $0.935$ \\
    0.5  & $1.62 \pm 0.09\times$          & $0.938$ \\
    1.0  & $1.61 \pm 0.06\times$          & $0.939$ \\
    2.0  & $\mathbf{1.57 \pm 0.09\times}$ & $\mathbf{0.940}$ \\
    5.0  & $1.59 \pm 0.01\times$          & $0.941$ \\
    \bottomrule
  \end{tabular}
\end{table}

\begin{table}[ht]
  \caption{Effect of graph embedding dimension $d_e$ ($\lambda_\text{graph}$ fixed at 1.0). Mean over 3 seeds.}
  \label{tab:gemb_sweep}
  \centering
  \small
  \begin{tabular}{ccc}
    \toprule
    $d_e$ & Degradation Ratio     & Aff.\ Acc      \\
    \midrule
    8     & $1.69\times$          & $0.939$ \\
    16    & $1.67\times$          & $0.939$ \\
    32    & $\mathbf{1.58\times}$ & $0.939$ \\
    64    & $1.65\times$          & $0.939$ \\
    \bottomrule
  \end{tabular}
\end{table}

$\lambda_\text{graph} = 0$ (no graph loss) yields aff.\ accuracy of only 0.46 and $2.53\times$ degradation, confirming the graph loss is essential. Once $\lambda_\text{graph} \geq 0.5$, degradation stabilizes within a $0.05\times$ band regardless of further increases, indicating the model is not sensitive to the exact value in this range. $d_e$ has negligible effect: all four values achieve the same Aff.\ Acc (0.939) and degradation differences remain within $0.11\times$.

\section{Monotonicity Constraint Ablation}
\label{app:monotonicity}

We ablate the monotonicity constraint on Craftax ($D{=}4$), comparing four conditions: (1) \textbf{Vanilla}: no affordance graph; (2) \textbf{Graph (no constraint)}: graph-conditioned model with oracle graph targets, no OR-operation enforced at inference; (3) \textbf{Graph-AR (no constraint)}: same but with auto-regressive predicted graph; (4) \textbf{\ours{} (constrained)}: full model with monotonicity OR-operation.

\begin{table}[ht]
  \caption{8-step degradation ratio (MSE at step 8 / MSE at step 1) on Craftax. Mean $\pm$ std over 3 seeds. Lower is better.}
  \label{tab:monotonicity}
  \centering
  \small
  \begin{tabular}{lcc}
    \toprule
    Condition                & Degradation Ratio                     & Notes                     \\
    \midrule
    Vanilla                  & \nval{$51.4 \pm 26.8\times$}          & no graph                  \\
    Graph (no constraint)    & \nval{$136.3 \pm 176.0\times$}        & oracle graph, no OR-op    \\
    Graph-AR (no constraint) & \nval{$137.1 \pm 177.0\times$}        & predicted graph, no OR-op \\
    \ours{} (constrained)    & \nval{$\mathbf{1.10 \pm 0.01\times}$} & full model                \\
    \bottomrule
  \end{tabular}
\end{table}

The monotonicity constraint is decisive: \ours{} holds at $1.10\times$ degradation across all three seeds (std $= 0.01$), while all unconstrained variants diverge. Notably, graph-conditioned models \emph{without} the constraint perform far worse than Vanilla ($136.3\times$ vs.\ $51.4\times$). Without the OR-operation, the graph predictor can deactivate previously discovered affordances, introducing distribution shift that compounds more severely than having no graph at all. The constraint eliminates this failure mode by construction.

\section{Compute Resources}
\label{app:compute}

All experiments were conducted on a single NVIDIA RTX 4090 GPU (24\,GB VRAM). No distributed or cloud compute was used.

\section{MiniHack LavaCross Details}
\label{app:minihack}

MiniHack-LavaCross-v0~\citep{samvelyan2021minihack} places the agent on a map separated from the goal by a lava corridor. A levitation potion is available nearby; drinking it activates the \texttt{lava\_traversable} affordance ($D{=}2$ SC chain: find potion $\to$ drink potion $\to$ cross lava). The environment has 77 discrete actions. Imagination MSE is computed in the glyph-embedding space (16-dim per glyph $\times$ 21$\times$79 grid) and is not directly comparable to pixel-space MSE in KeyDungeon/Crafter. Over 3 seeds, \ours{} achieves $1021.2\pm102.4 \times 10^{-3}$ MSE vs.\ Vanilla's $1496.2\pm1116.5 \times 10^{-3}$ ($1.47\times$ improvement). Notably, \ours{} exhibits substantially lower cross-seed variance (std $102.4$) than Vanilla (std $1116.5$), indicating that affordance graph conditioning stabilizes imagination quality across random initializations.

\newpage

\section*{Usage of Large Language Models}
During the preparation of this manuscript, large language models were used solely for grammar checking and writing refinement. They were not used as any component of the AGWM method, did not contribute to experimental design or result analysis, and had no impact on the scientific content or conclusions of this paper.

\end{document}